\documentclass{article}
\usepackage{spconf,amsmath,graphicx}
\usepackage{times,amsmath,epsfig}
\usepackage{bm}
\usepackage{amsfonts}
\usepackage[numbers,sort&compress]{natbib}
\emergencystretch=\maxdimen
\begin{document}\sloppy
\def\x{{\mathbf x}}
\def\L{{\cal L}}

\title{Mixed-Resolution Image Representation and\\ Compression with Convolutional Neural Networks}
%
\name{Lijun Zhao{\small $~^{\star}$},Huihui Bai{\small $~^{\star}$}, Feng Li{\small $~^{\star}$}, Anhong Wang{\small $~^{\#}$}, Yao Zhao{\small $~^{\star}$}
\thanks{Corresponding author: Huihui Bai (email: hhbai@bjtu.edu.cn). This work was supported in part by National Natural Science Foundation of China (No. 61672087, 61672373) and the Fundamental Research Funds for the Central Universities (K17JB00260, K17JB00350).}}

\address{{\small $~^{\star}$}Institute Information Science, Beijing Jiaotong University, P. R. China\\
{\small $~^{\#}$}Institute of Digital Media and Communication, Taiyuan University of Science and Technology}

%
%
%

%
\maketitle
\begin{abstract}
In this paper, we propose an end-to-end mixed-resolution image compression framework with convolutional neural networks. Firstly, given one input image, feature description neural network (FDNN) is used to generate a new representation of this image, so that this image representation can be more efficiently compressed by standard codec, as compared to the input image. Furthermore, we use post-processing neural network (PPNN) to remove the coding artifacts caused by quantization of codec. Secondly, low-resolution image representation is adopted for high efficiency compression in terms of most of bit spent by image's structures under low bit-rate. However, more bits should be assigned to image details in the high-resolution, when most of structures have been kept after compression at the high bit-rate. This comes from a fact that the low-resolution image representation can't burden more information than high-resolution representation beyond a certain bit-rate. Finally, to resolve the problem of error back-propagation from the PPNN network to the FDNN network, we introduce to learn a virtual codec neural network to imitate two continuous procedures of standard compression and post-processing. The objective experimental results have demonstrated the proposed method has a large margin improvement, when comparing with several state-of-the-art approaches.
\end{abstract}
\section{Introduction}

Image and video data appearing as general media provides us great convenience to share information and communicate with each other. However, nowadays huge amounts of image and video are required to be stored and transmitted efficiently. As we all know, image and video coding techniques have enormously alleviated this problem by compressing these data to be a small yet strong expressive one, but the compression efficiency of standard coding gradually can't satisfy the explosive transmission demands of social media and streaming media with the popularization of electronic products such as digital camera and cell-phone \cite{p1}. Thus, image and video’s representation as well as compression towards higher compression ratio should be deeply studied, especially using deep learning.

Conventional still image coding \cite{p2,p3} has been developed from JPEG and JPEG2000 to WebP and BPG, etc. Meanwhile, several latest works, such as \cite{p4,p5,p6,p7,p8}, are devoted to image compression with deep neural networks. In \cite{p6}, two collaborated convolutional neural networks are used to form a unified end-to-end learning framework, where one network produces a compact representation for encoding, while another one reconstructs the decoded image. Different the works of \cite{p6}, a virtual codec neural network of \cite{p7} is learned to bridge the gap between the networks ahead of standard codec and after this codec so that the gradients could be properly passed from the back end to the front end. On the basis of the works of \cite{p7}, multiple description convolutional neural networks are designed to compress image so as to ensure that an acceptable image can be decoded in the un-prioritized network or under the condition of transmission congestion \cite{p8}.

Because our work is highly related to the problem of post-processing such as artifacts removal \cite{p9}, de-blocking \cite{p10} and de-noising \cite{p11}, we next introduce several state-of-the-art works \cite{p11,p12,p13,p14,p15,p16,p17} for compression artifacts removal. In \cite{p11}, shape-adaptive discrete cosine transform-based filtering is developed for de-noising and de-blocking by introducing the shape of the transform's support in a point-wise adaptive fashion. In \cite{p12}, after grouping similar 2D image patches into 3D data arrays, three successive procedures of 3D transformation of these array, shrinkage of the transform spectrum, and inverse 3D transformation are conducted to achieve image de-noising. In \cite{p13}, two-step algorithm is formed to reducing artifacts by dictionary learning and total variation regularization. To reduce blocking artifacts and obtain high-quality image, an optimization problem using constrained non-convex low-rank model is developed within maximum a posteriori framework \cite{p14}. Except the methods of filtering and optimization \cite{p11,p12,p13,p14}, there are several convolutional neural network-based approaches, such as \cite{p15,p16,p17}. Although post-processing could improve the coding efficiency, they lose the sight of the significance of image representation, which can highlight the significance pixels or regions before coding in order to protect these pixels or regions. Thus, post-processing and image representation should be combined together to further improve image's coding efficiency.

Although the literatures of \cite{p6, p7} have got a high image compression gains at the low bit-rate, they can't have a large margin coding gain at the high bit-rate. In this paper, mixed-resolution image representation and compression with deep convolutional neural networks is introduced to efficiently compress image, no matter which bit-rate is chosen by the users. The rest of this paper is arranged as follow. Firstly, we introduce the proposed method in Section 2.  Secondly, the experimental results are given in Section3. At last, the Section 4 concludes the paper.

\section{The proposed method}

Our framework is composed of feature description neural network (denoted as FDNN) network, a standard codec (e.g., JPEG), post-processing neural network (PPNN), and virtual codec neural network (VCNN) network. To greatly reduce the amount of image data for storage or transmission, we use the FDNN network to represent ground-truth image $\bm{X} \in \mathbb{R}^{M\times N}$ as $\bm{Y}$ in the low-resolution or high-resolution before image compression. For simplicity, the FDNN network is expressed as a non-linear function $f(\bm{X},\alpha)$, in which $\alpha$ is the parameter set of FDNN network. The compression procedure of standard codec is described as a mapping function $\bm{Z}=g(\bm{Y},\beta)$, where $\beta$ is the parameter set of codec. Our PPNN network learns a post-processing function $h(\bm{Z},\gamma)$ from image $\bm{Z}$ to image $\bm{X}$ to remove the noise, such as blocking artifacts, ringing artifacts and blurring, when $\bm{Y}$ is represented in the high resolution. Here, the parameter $\gamma$ is the parameter set of PPNN network. However, if $\bm{X}$ is represented in the low-resolution, PPNN network is used to simultaneously de-artifact and up-sample $\bm{Z}$ to be $\bm{\tilde{I}}$ from low-resolution yet low-quality to high-resolution yet high quality.

To combine image compression standard with image representation as well as post-processing based on convolutional neural network, the intuitive idea is that the compression procedure of codec is learned by a convolutional neural network to get a approximation function. Although convolutional neural network is a powerful tool to approximate any nonlinear function, it's hard to imitate the procedure of image compression, because the quantization operator always leads to serious blocking artifacts and coding distortion. However, as compared to the compressed images $\bm{Z}$, the post-processed compressed image $\bm{\tilde{I}}$ has less distortion, because $\bm{\tilde{I}}$ loses some detail information, but it does not have obvious artifacts and blocking artifacts. Thus, the function $h(g(\bm{Y},\beta),\gamma)$ of two successive procedure of codec $g(\bm{Y},\beta)$ and post-processing $h(\bm{Z},\gamma)$ can be well represented by the VCNN network. To make sure that the gradient can be rightly back-propagated from the PPNN to FDNN, our VCNN network is proposed to learn a projection function $v(\bm{Y},\theta)$ from image representation $\bm{Y}$ to final output $\bm{\tilde{I}}$ of PPNN. Here, the parameter $\theta$ is the parameter set of VCNN network. This projection can rightly approximate the two successive procedures: the compression of standard codec and post-processing based on convolutional neural network. After training the VCNN network, this network is leveraged to supervise the training of our FDNN network.

\subsection{Objective function}
Our framework's objective function is written as follows:
\begin{equation}
\begin{split}
\mathop{\arg\min}_{\alpha, \gamma, \theta} L(\bm{X},\bm{\tilde{I}})+ L(\bm{\hat{I}},\bm{\tilde{I}})+L_{SSIM}(s(\bm{Y}),\bm{X}), \\
\bm{Y}=f(\bm{X},\alpha),\bm{\tilde{I}}=h(\bm{Z},\gamma), \bm{Z}=g(\bm{Y},\beta), \bm{\hat{I}}=v(\bm{Y},\theta),
\end{split}
\end{equation}
where $\alpha$, $\gamma$, and $\theta$ are respectively three parameter sets of FDNN, PPNN, and VCNN network, and $s(\cdot)$ is the linear up-sampling operator, if $\bm{Y}$ and $\bm{X}$ don't have the same image size, or else $s(\bm{Y})=\bm{Y}$. Here, in order to make final output image $\bm{\tilde{I}}$ to be similar to $\bm{X}$, $L(\bm{X},\bm{\tilde{I}})$ has the L1 content loss $L_{content}(\bm{X},\bm{\tilde{I}})$ and L1 gradient difference loss $L_{gradient}(\bm{X},\bm{\tilde{I}})$ for the regularization of the FDNN network's training, which are written as::
\begin{equation}
\begin{split}
L_{content}(\bm{X},\bm{\tilde{I}})= \frac{1}{M \cdot N}\sum_{i}(||\bm{X}_i-\bm{\tilde{I}}_i||_1),
\end{split}
 \label{eqn::data loss}
\end{equation}

\begin{equation}
\begin{split}
L_{gradient}(\bm{X},\bm{\tilde{I}})= \frac{1}{M \cdot N} \sum_{i} (\sum_{k\in{\Omega}}||\nabla_k \bm{X}_i)-\nabla_k \bm{\tilde{I}}_i||_1)
\end{split}
\label{eqn::GD loss}
\end{equation}
where $||\cdot||_1$ is the L1 norm, which has better performance to supervise convolutional neural network's training than the L2 norm. This has been reported in the literature of \cite{p21}, which learns to predict subsequent frames from the video sequences.

Since standard codec, as a big obstacle, exists between PPNN network and FDNN network, it's tough to make the gradient back-propagate between them. Therefore, it's a challenging task to train FDNN network directly without the supervision of PPNN network. To address this task, we can learn a nonlinear function from the $\bm{Y}$ to $\bm{\tilde{I}}$ in the VCNN network, where the L1 content loss $L_{content}(\bm{\hat{I}},\bm{\tilde{I}})$  and L1 gradient difference loss $L_{gradient}(\bm{\hat{I}},\bm{\tilde{I}})$ are used to supervise the VCNN network's training. Here, $\bm{\hat{I}}$ is the predicted result by VCNN network to approximate $\bm{\tilde{I}}$.

The structural information of representation $\bm{Y}$ is always expected to be similar to ground-truth image $\bm{X}$, so the SSIM loss $L_{SSIM}(s(\bm{Y}),\bm{X})$ \cite{p22} supervises the learning of FDNN, besides the training loss from the network of VCNN, which is defined as follows:
\begin{equation}
\begin{split}
L_{SSIM}(s(\bm{Y}),\bm{X})=1-\frac{1}{M \cdot N} \sum_{i} L_{SSIM}(s(\bm{Y})_i,\bm{X}_i)
\end{split}
\label{eqn::SSIMLOSS}
\end{equation}

\begin{align}
&L_{SSIM}(s(\bm{Y})_i,\bm{X}_i)=\notag\\
&\frac{(2\mu_{s(\bm{Y})_i}\cdot \mu_{\bm{X}_i}+c1)(2\sigma_{s(\bm{Y})_i \bm{X}_i}+c2)}{(\mu^2_{s(\bm{Y})_i}+\mu^2_{\bm{X}_i}+c1)(\sigma^2_{s(\bm{Y})_i}+\sigma^2_{\bm{X}_i}+c2)}
\end{align}
where $c1$ and $c2$ are two constant values, which respectively equal to $0.0001$ and $0.0009$. $\mu_{\bm{X}_i}$ and $\sigma^2_{\bm{X}_i}$ respectively denote the mean value and the variance of the neighborhood window centered by pixel $i$ in the image $\bm{X}$. In this way, $\mu_{s(\bm{Y})_i}$ as well as $\sigma^2_{s(\bm{Y})_i}$ can be denoted similarly. Meanwhile, $\sigma_{s(\bm{Y})_i \bm{X}_i}$ is the covariance between neighbourhood windows centered by pixel $i$ in the image $\bm{X}$ and in the image $s(\bm{Y})$. Because the function of SSIM is differentiable, the gradient can be efficiently back-propagated during the FDNN network's training.

\subsection{Networks}
Eight convolutional layers in the FDNN network are used to extract features so as to represent the ground-truth image $\bm{X}$ as $\bm{Y}$. In this network, the weights of these convolutional layers are in the spatial size of 9x9 for the first layer and the last layer, which could make receptive field (RF) of convolutional neural networks to be large enough. In addition, other six convolutional layers in the FDNN use 3x3 convolution kernel to further enlarge the size of RF. These convolutional layers are used to increase the nonlinearity of the network, when ReLU is followed to activate the output features of these convolutional hidden layers. The feature map number of 1-7 convolutional layers is 128, but the last layer only has one feature map so as to keep consistent with the ground truth image $\bm{X}$. Each convolutional layer is operated with a stride of 1, except that the second layer uses stride step of 2 to down-sample feature maps, so that the convolution operation is carried out in the low-resolution space to reduce computational complexity from the third convolutional layer to the 8-th convolutional layer. However, the second layer uses stride step of 1, if $\bm{Y}$ is represented in the high-resolution, when the given bit-rate is beyond a certain value. All the convolutional layers are followed by an activation layer with ReLU function, except the last convolutional layer.

In the PPNN network, we leverage seven convolutional layers to extract features and each layer is activated by ReLU function. The size of convolutional layer is 9x9 in the first layer and the left six layers use 3x3, while the output channel of feature map equals to 128 in these convolutional layer. After these layers, one de-convolution layer with size of 9x9 and stride to be 2 is used to up-scale feature map from low-resolution to high-resolution so that the size of output image is matched with the ground truth image. However, if $\bm{Y}$ is full-resolution image, the last de-convolution layer is replaced by convolutional layer with size of 9x9 and stride to be 1.

The VCNN network is designed to be the same structure with the PPNN network, because they are the same kind of low-level image processing problems. The role of VCNN network is to make the representation $\bm{Y}$ degrade to a post-processed compressed but high-resolution image $\bm{\tilde{I}}$. On the contrary, the functionality of the PPNN network is to improve the quality of the compressed represenation $\bm{Z}$ so that the user could receive a high-quality image $\bm{\tilde{I}}$ without coding artifacts after post-processing with PPNN network at the decoder.
\subsection{Learning Algorithm}
Due to the difficulty of directly training the whole framework once, we decompose the learning of three convolutional neural networks in our framework as three sub-problems learning. First, we initialize all the parameter set $\beta$, $\alpha$, $\gamma$, and $\theta$ of codec, FDNN network, PPNN network, and VCNN network. Meanwhile, we use traditional interpolation methods to get an initial representation image $\bm{Y}$ of the ground-truth image $\bm{X}$, which is then compressed by JPEG codec as the input of training data set at the beginning. Next, the first sub-problem learning is to train PPNN network by updating the parameter set of $\gamma$ according to the Eq. (2-3). The compressed representation image $\bm{Z}$ got from ground-truth image $\bm{X}$ and its post-processed compressed image $\bm{\tilde{I}}$ predicted by PPNN network are used for the second sub-problem's learning of VCNN to update parameter set of $\theta$ based on the Eq. (4-5). After VCNN's learning, we fix the parameter set of $\theta$ in the VCNN network to carry on the third sub-problem learning by updating the parameter set of $\alpha$ for training FDNN network. After FDNN network's learning, the next iteration begins to train the PPNN network, after the updated description image are compressed by the standard codec. It is worth mentioning that the functionality of VCNN network is used to bridge the great gap between FDNN and PPNN. Thus, once the training of our whole framework is finished, the VCNN network is not in use any more, that is to say, only the parameter sets of $\alpha$, $\gamma$ in the networks of FDNN and PPNN are employed during testing.

\section{Experimental results}
\begin{figure}[t]
\centering
\includegraphics[width=3.0in]{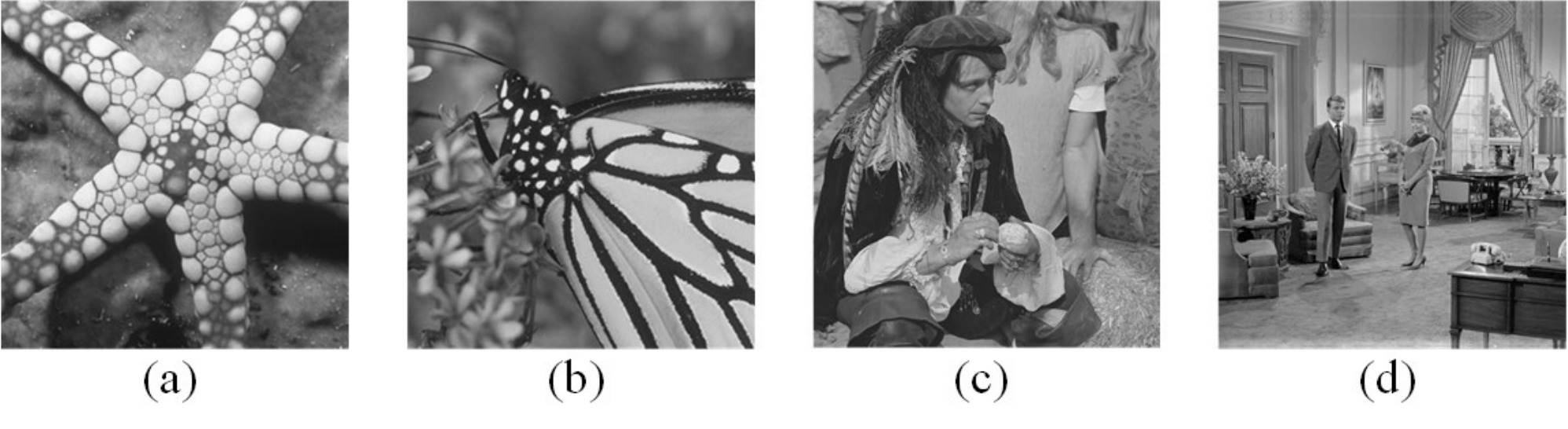}
\caption{The data-set is used for our testing}
\label{Fig3}
\end{figure}
\subsection{Training details}
Our framework of learning a virtual codec neural network to compress image is implemented with TensorFlow \cite{p20}. The training data comes from \cite{p19}, in which 400 images of size 180x180 are included. We augment these data by cropping, rotating and flipping image to build our training data-set, in which the total number of image patches with size of 160x160 is 3200 (n=3200). For testing as shown in Fig. \ref{Fig3}, four images, which are broadly employed for compressed image de-noising or de-artifact, are used to evaluate the efficiency of the proposed method. We train our model using the optimization method of Adam, with the beta1=0.9, beta2=0.999. The initial learning rate of training three convolutional neural network is set to be 0.0001, while the learning rate decays to be half of the initial one once the training step reaches 3/5 of total step. And it decreases to be 1/4 of the initial one when the training step reaches 4/5 of total step.

\subsection{The quality comparison of different methods}
To validate the efficiency of the proposed framework, we compare our method with JPEG, Foi's \cite{p11}, BM3D \cite{p12}, DicTV \cite{p13}, CONCOLOR \cite{p14}, and Jiang's \cite{p6}. Here, both Foi's \cite{p11} and BM3D \cite{p12} are the class of image de-noising. The method of Foi's \cite{p11} is specifically designed for de-blocking. The approaches of DicTV \cite{p13}, CONCOLOR \cite{p14} use the dictionary learning or the low-rank model to resolve the problem of de-blocking and de-artifact. The results of Foi's \cite{p11}, BM3D \cite{p12}, DicTV \cite{p13} and CONCOLOR \cite{p14} are got by strictly using the author's open codes with the parameter settings in their papers. However, the highly related method of Jiang's \cite{p6} only give one factor for testing, so we try to re-implement their method with TensorFlow. Meanwhile, to fairly compare with the Jiang's \cite{p6}, we use our FDNN and PPNN to replace networks of ComCNN and ReCNN for training and testing to avoid the effect of the network's structure design on the experimental results. Besides, we extend Jiang's framework \cite{p6} to be mixed-resolution so that more comparisons can be conducted between the proposed method and Jiang's \cite{p6}.
\begin{figure}[t]
\centering
\includegraphics[width=3.5in]{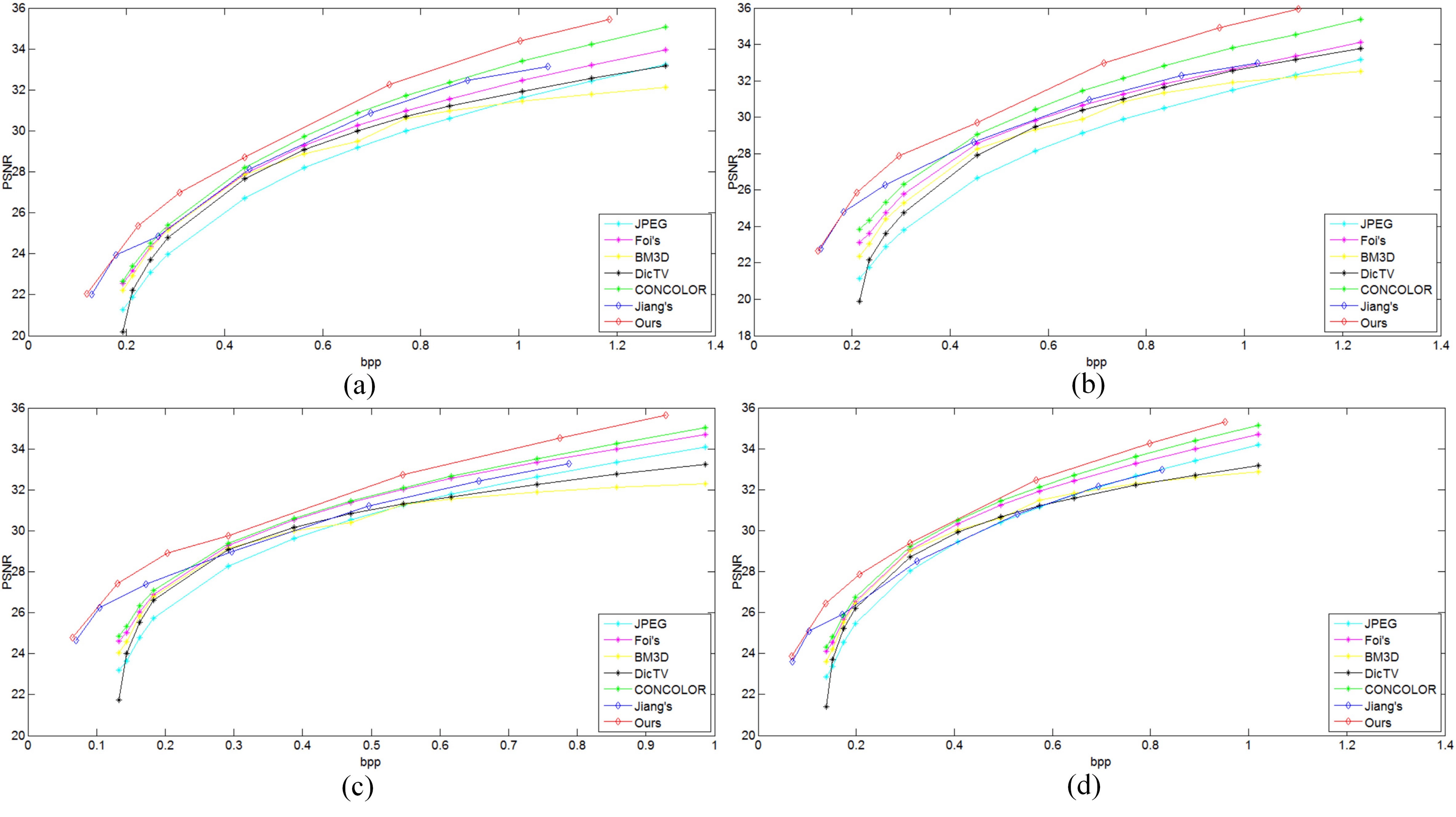}
\caption{The objective measurement comparison on PSNR for several state-of-the-art approaches. (a-d) are the results of image (a-d) in Fig. \ref{Fig3}}
\label{Fig4}
\end{figure}
The JPEG software of image compression in OpenCV is used for all the experimental results. To compare our method with Jiang's \cite{p6} in the following, the low resolution representation are compressed by JPEG with quality factors to be 5, 10, and 20 at low bit-rate, but the high resolution representation is assigned with 10, 20, 30, and 40 for their training and testing at high bit-rate. Meanwhile, Foi's \cite{p11}, BM3D \cite{p12}, DicTV \cite{p13}, and CONCOLOR \cite{p14} compress image with JPEG codec with quality factors to be 2, 3, 4, 5, 10, 15, 20, 25 and 30. Note that in the proposed framework the JPEG codec is used, but in fact our framework can be applied into most of existing standard codec.

We use the Peak Signal to Noise Ratio (PSNR) as the objective quality's measurement. From the Fig. \ref{Fig4}, where bpp denotes the bit-per-pixel, it can be obviously observed that the proposed method has the best objective performance on PSNR and SSIM, as compared to several state-of-the-art approaches: JPEG \cite{p2}, Foi's \cite{p11}, BM3D \cite{p12}, DicTV \cite{p13}, CONCOLOR \cite{p14}, and Jiang's \cite{p6}. From the above comparisons, it can be known that the back-propagation of gradient in the feature description network from postprocessing neural network plays a significant role on the effectiveness of feature description and the compression efficiency when combining the neural network with standard codec together to effectively compress image. In a word, by learning a virtual codec neural network, the proposed framework provides a good way to resolve the gradient back-propagation problem in the image compression framework with convolutional neural network ahead of a standard codec.

\section{Conclusion}
In this paper, we propose an end-to-end mixed-resolution image compression framework to resolve the problem of non-differentiability of the quantization function in the lossy image compression by learning a virtual codec neural network. Directly learning the whole framework of the proposed method is a intractable problem, so we decompose this challenging optimization problem into three sub-problems learning. Finally, experimental results have shown the priority of the proposed method than several state-of-the-art methods.

\bibliographystyle{IEEEbib}
\bibliography{Template}

\begin{thebibliography}{10}

\bibitem{p1}
W.~Dai, G.~Cheung, N.~Cheung, A.~Ortega, and O.~Au,
\newblock ``{Merge frame design for video stream switching using piecewise
  constant functions},''
\newblock {\em IEEE Transactions on Image Processing}, vol. 25, no. 8, pp.
  3489--3504, 2016.

\bibitem{p2}
G.~Wallace,
\newblock ``{The JPEG still picture compression standard},''
\newblock {\em IEEE Transactions on Consumer Electronics}, vol. 38, no. 1, pp.
  xviii--xxxiv, 1992.

\bibitem{p3}
J.~Lainema, M.~Hannuksela, V.~Vadakital, and E.~Aksu,
\newblock ``{HEVC still image coding and high efficiency image file format},''
\newblock in {\em IEEE International Conference on Image Processing}, Arizona,
  Sept. 2016.

\bibitem{p4}
G.~Toderici, D.~Vincent, N.~Johnston H., Jin, D.~Minnen, J.~Shor, and
  M.~Covell,
\newblock ``{Full resolution image compression with recurrent neural
  networks},''
\newblock in {\em IEEE Conference on Computer Vision and Pattern Recognition},
  Honolulu, July 2017.

\bibitem{p5}
M.~Li, W.~Zuo, S~Gu., D.~Zhao, and D.~Zhang,
\newblock ``Learning convolutional networks for content-weighted image
  compression,''
\newblock in {\em arXiv: 1703.10553}, 2017.

\bibitem{p6}
F.~Jiang, W.~Tao, S.~Liu, J.~Ren, X.~Guo, and D.~Zhao,
\newblock ``An end-to-end compression framework based on convolutional neural
  networks,''
\newblock {\em IEEE Transactions on Circuits and Systems for Video Technology},
  2017.

\bibitem{p7}
L.~Zhao, H.~Bai, A.~Wang, and Y.~Zhao,
\newblock ``Learning a virtual codec based on deep convolutional neural network
  to compress image,''
\newblock in {\em arXiv: 1712.05969}, 2017.

\bibitem{p8}
L.~Zhao, H.~Bai, A.~Wang, and Y.~Zhao,
\newblock ``Multiple description convolutional neural networks for image
  compression,''
\newblock in {\em arXiv: 1801.06611}, 2018.

\bibitem{p9}
L.~Zhao, H.~Bai, A.~Wang, Y.~Zhao, and B.~Zeng,
\newblock ``Two-stage filtering of compressed depth images with markov random
  field,''
\newblock {\em Signal Processing: Image Communication}, vol. 51, pp. 11--22,
  2017.

\bibitem{p10}
P.~List, A.~Joch, J.~Lainema, G.~Bjontegaard, and M.~Karczewicz,
\newblock ``Adaptive deblocking filter,''
\newblock {\em IEEE Transactions on Circuits and Systems for Video Technology},
  vol. 13, no. 7, pp. 614--619, 2003.

\bibitem{p11}
A.~Foi, V.~Katkovnik, and K.~Egiazarian,
\newblock ``{Pointwise shape-adaptive DCT for high-quality denoising and
  deblocking of grayscale and color images},''
\newblock {\em IEEE Transactions on Image Processing}, vol. 16, no. 5, pp.
  1395--1411, 2007.

\bibitem{p12}
K.~Dabov, A.~Foi, V.~Katkovnik, and K.~Egiazarian,
\newblock ``{Image denoising by sparse 3-D transform-domain collaborative
  filtering},''
\newblock {\em IEEE Transactions on Image Processing}, vol. 16, no. 8, pp.
  2080--2095, 2007.

\bibitem{p13}
H.~Chang, M.~Ng, and T.~Zeng,
\newblock ``{Reducing artifacts in JPEG decompression via a learned
  dictionary},''
\newblock {\em IEEE Signal Processing Letters}, vol. 62, no. 3, pp. 718--728,
  2014.

\bibitem{p14}
J.~Zhang, R.~Xiong, C.~Zhao, Y.~Zhang, S.~Ma, and W.~Gao,
\newblock ``{CONCOLOR: Constrained non-convex low-rank model for image
  deblocking},''
\newblock {\em IEEE Transactions on Image Processing}, vol. 25, no. 3, pp.
  1246--1259, 2016.

\bibitem{p15}
C.~Dong, Y.~Deng, C.~Chen, and X.~Tang,
\newblock ``{Compression Artifacts Reduction by a Deep Convolutional
  Network},''
\newblock in {\em IEEE International Conference on Computer Vision}, Santiago,
  Dec. 2015.

\bibitem{p16}
L.~Cavigelli, P.~Hager, and L.~Benini,
\newblock ``{CAS-CNN: A deep convolutional neural network for image compression
  artifact suppression},''
\newblock in {\em IEEE Conference on Neural Networks}, Anchorage, AK, USA, May
  2017.

\bibitem{p17}
L.~Galteri, L.~Seidenari, M.~Bertini, and B.~Del,
\newblock ``Deep generative adversarial compression artifact removal,''
\newblock in {\em arXiv: 1704.02518}, 2017.

\bibitem{p21}
M.~Mathieu, C.~Couprie, and Y.~LeCun,
\newblock ``Deep multi-scale video prediction beyond mean square error,''
\newblock in {\em arXiv: 1511.05440}, 2015.

\bibitem{p22}
Z.~Wang, A.~Bovik, H.~Sheikh, and E.~Simoncelli,
\newblock ``Image quality assessment: from error visibility to structural
  similarity,''
\newblock {\em IEEE Transactions on Image Processing}, vol. 13, no. 4, pp.
  600--612, 2004.

\bibitem{p20}
M.~Abadi, A.~Agarwal, P.~Barham, E.~Brevdo, Z.~Chen, C.~Citro, and et~al.,
\newblock ``Tensorflow: large-scale machine learning on heterogeneous
  distributed systems,''
\newblock in {\em arXiv:1603.04467}, 2016.

\bibitem{p19}
Y.~Chen and T.~Pock,
\newblock ``Trainable nonlinear reaction diffusion: A flexible framework for
  fast and effective image restoration,''
\newblock {\em IEEE Transactions on Pattern Analysis and Machine Intelligence},
  vol. 39, no. 6, pp. 1256--1272, 2017.

\end{thebibliography}

\end{document}